\relax
\documentclass[11pt,a4paper]{article} %
\usepackage[hyperref]{acl2020} 

\usepackage{times}
\usepackage{soul}
\usepackage{url}
\usepackage[utf8]{inputenc}
\usepackage{graphicx}
\usepackage{amsmath}
\usepackage{amsthm}
\usepackage{booktabs}
\usepackage{algorithm}
\usepackage{algorithmic}
\usepackage{arydshln}
\usepackage{amsmath} 
\usepackage{booktabs} 
\usepackage{multirow} 
\usepackage{makecell} 
\usepackage{comment} 
\usepackage[normalem]{ulem} 
\usepackage{blindtext, subfig} 
\usepackage{amsbsy} 
\usepackage{bm} 
\usepackage{array}
\usepackage{hhline}
\usepackage{tabularx} 

\usepackage{color}
\usepackage{enumitem}
\usepackage[linecolor=red]{todonotes}

\definecolor{red}{RGB}{189, 26, 26}
\definecolor{green}{RGB}{228, 249, 189}
\definecolor{Green}{RGB}{141, 213, 134}
\definecolor{purple}{RGB}{235, 219, 246}
\definecolor{pink}{RGB}{250, 221, 236}
\definecolor{smt}{RGB}{195, 62, 0}
\definecolor{nmt}{RGB}{252, 133, 0}
\definecolor{gmt}{RGB}{0, 98, 51}
\definecolor{pipe}{RGB}{0,128,255}

\usepackage{amsmath,booktabs,varwidth,collcell}
\usepackage{ulem}

\newcommand{\fixedwidthcolumn}[1]{%
  \begin{varwidth}[t]{\fcolwidth}%
    \raggedright
    \strut#1\strut
  \end{varwidth}}

\newcolumntype{L}[1]{>{\gdef\fcolwidth{#1}\collectcell\fixedwidthcolumn}l<{\endcollectcell}}

\newcommand{\pnt}[1]{{\scriptstyle#1}} 

\DeclareUrlCommand{\bulurl}{} 
\usepackage{microtype}

\aclfinalcopy 
\def\aclpaperid{323} 

\title{\textit{Cue Me In}: Content-Inducing Approaches to Interactive Story Generation}

\author{Faeze Brahman$^\dagger$, Alexandru Petrusca$^\dagger$, and Snigdha Chaturvedi$^\ddagger$\\
$^\dagger$Department of Computer Science and Engineering, University of California, Santa Cruz\\
$^\ddagger$Department of Computer Science, University of North Carolina at Chapel Hill\\
\texttt{\{fbrahman, apetrusc\}@ucsc.edu} \qquad \texttt{snigdha@cs.unc.edu} \\}

\begin{document}

\maketitle

\begin{abstract}
    Automatically generating stories is a challenging problem that requires producing causally related and logical sequences of events about a topic. Previous approaches in this domain have focused largely on one-shot generation, where a language model outputs a complete story based on limited initial input from a user. Here, we instead focus on the task of interactive story generation, where the user provides the model mid-level sentence abstractions in the form of cue phrases \textit{during} the generation process. This provides an interface for human users to guide the story generation. We present two content-inducing approaches to effectively incorporate this additional information. Experimental results from both automatic and human evaluations show that these methods produce more topically coherent and personalized stories compared to baseline methods. 
\end{abstract}

\section{Introduction}
Automatic story generation requires composing a coherent and fluent passage of text about a sequence of events. Prior studies on story generation mostly focused on symbolic planning ~\cite{Lebowitz:87,Perez:01,Porteous:09,Reidl:10} or case-based reasoning ~\cite{Gervas:05} that heavily relied on manual knowledge engineering. \par 
	\begin{figure}[!hbt]
		\begin{center}
		\includegraphics[width=0.9\columnwidth]{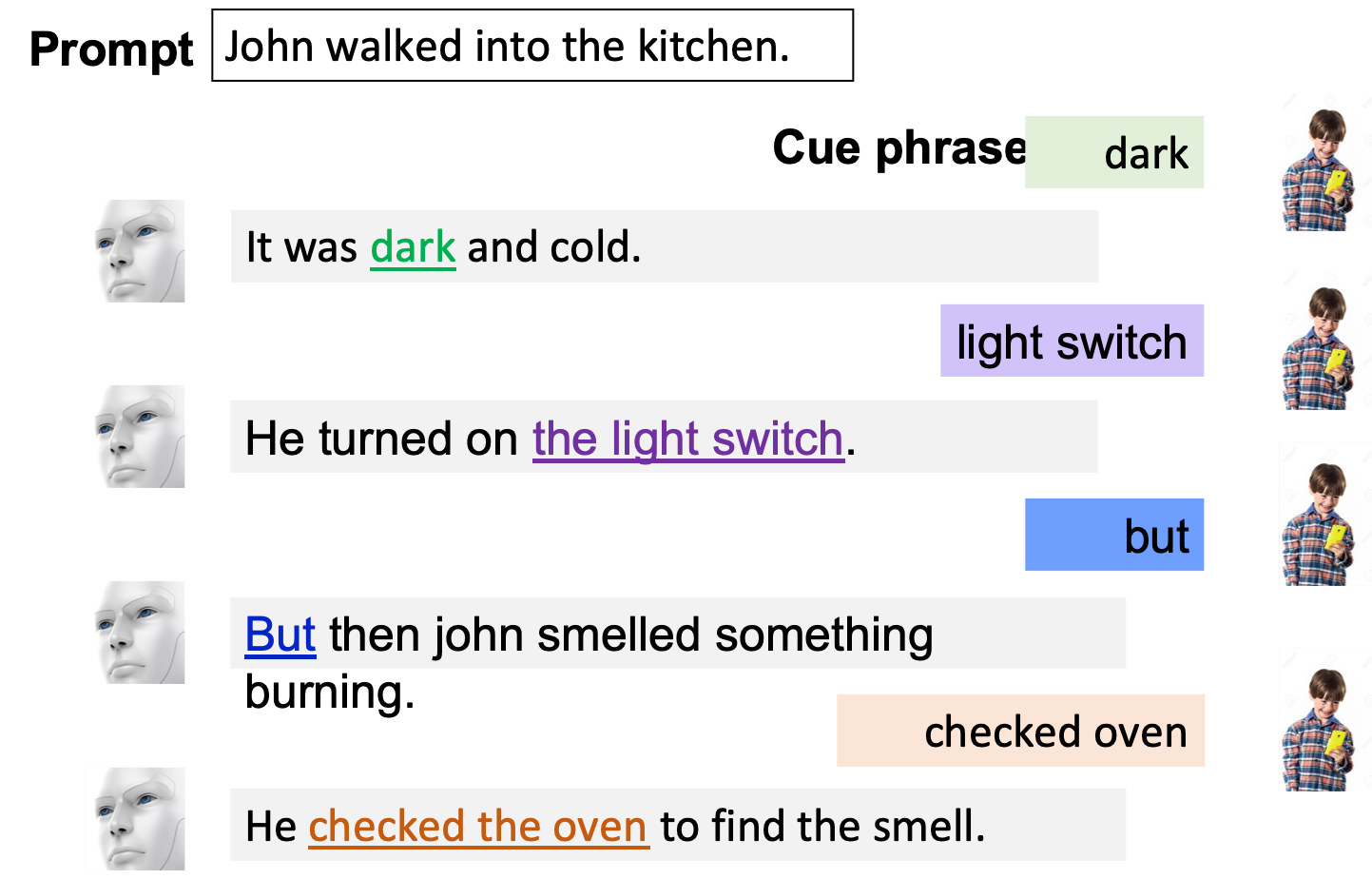}
		\caption{Interactive story generation: the user inputs the first sentence of the story (prompt), and provides guiding cue phrases as the system generates the story one sentence at a time.} 
		\label{fig: problem}
		\end{center}
	\end{figure}

Recent state-of-the-art methods for story generation ~\cite{Martin:18,Clark:18} are based on sequence-to-sequence models ~\cite{sutskever:14:10} that generate a story in one go. In this setting, the user has little control over the generated story.

On the other hand, when humans write, they incrementally edit and refine the text they produce. Motivated by this, rather than generating the entire story at once,  we explore the problem of interactive story generation.
In this setup, a user can provide the model mid-level sentence abstractions in the form of \emph{cue phrases} as the story is being generated. Cue phrases enable the user to inform the system of what they want to happen next in the story and have more control over what is being generated. To achieve our goal, this paper primarily focuses on approaches for smoothly and effectively incorporating user-provided cues.
The schematic in Fig.~\ref{fig: problem} illustrates this scenario: the system generates the story one sentence at a time, and the user guides the content of the next sentence using cue phrases. We note that the generated sentences need to fit the context, and also be semantically related to the provided cue phrase. \par

A fundamental advantage of using this framework as opposed to a fully automated one is that it can provide an interactive interface for human users to incrementally supervise the generation by giving signals to the model throughout the story generation process. This human-computer collaboration can result in generating richer and personalized stories. In particular, this field of research can be used in addressing the literacy needs of learners with disabilities and enabling children to explore creative writing at an early age by crafting their own stories. 

In this paper, we present two content-inducing approaches based on the Transformer Network~\cite{Vaswani:17} for interactively incorporating 
external knowledge 
when automatically generating stories. Here, our external knowledge is in the form of cue phrases provided by the user to enable interaction, but can readily be replaced 
with knowledge accessible through other means\footnote{For example, the user-provided cues can be replaced by the outputs of an automatic planner. Our models are flexible enough to work in other setups.}. Specifically, our models fuse information from the story context and cue phrases through a hierarchical attention mechanism. The first approach, \textit{Cued Writer}, employs two independent encoders (for incorporating context and cue phrases) and an additional attention component to capture the semantic agreement between the cue phrase and output sentence. The second approach,  \textit{Relevance Cued Writer}, additionally measures the relatedness between the context and cue phrase through a context-cue multi-head unit. In both cases, we introduce different attention units in a single end-to-end neural network. \par 

Our automatic and human evaluations demonstrate that the presented models outperform strong baselines and can successfully incorporate cues in generated stories. This capability is one step closer to an interactive setup, and unlike one-shot generation, it lets users have more control over the generation.
Our contributions are twofold:
\begin{itemize} [noitemsep,nolistsep]
    \item Two novel content-inducing approaches to incorporate additional information, in this case cue phrases, into the generation phase.
    \item Experiments demonstrating utility of content-inducing approaches using automatic and human evaluations.
\end{itemize}

\section{Related Work}
Automatic story generation is a longstanding problem in AI, with early work dating back to the 1970s based on symbolic planning~\cite{Lebowitz:87,Perez:01,Porteous:09,Reidl:10} and case-based reasoning using ontologies~\cite{Gervas:05}.
~\citet{crowdsource} extended prior works toward learning domain models (via corpus and/or crowdsourcing) to support open story generation about any topic.\par 
With the advent of deep learning there has been a major shift towards using seq2seq models~\cite{sutskever:14:10,bahdanau:14} for various text generation tasks, including 
storytelling~\cite{roemmele2016writing,jian:17, hu20aaai}. However, these models often fail to ensure coherence in the generated story. To address this problem, \citeauthor{Clark:18} \shortcite{Clark:18} incorporated entities given their vector representations, which get updated as the story unfolds. Similarly, \citet{char-centric} proposed a character-centric story generation by learning character embeddings directly from the corpus.~\citeauthor{Fan:18} \shortcite{Fan:18} followed a two-step process to first generate the premise and then condition on that to generate the story. \citet{DBLP:conf/aaai/YuLLTZ0020} proposed a multi-pass CVAE to improve wording diversity and content consistency.

Previous work has explored the potential of creative writing with a machine in the loop. \citeauthor{clarkCreative} \shortcite{clarkCreative} found that people generally enjoy collaborating with a machine. Traditional methods proposed to write stories collaboratively using a case-based reasoning architecture~\cite{swanson_say_2012}. Recent work~\cite{roemmele_creative_2015} extended this to find relevant suggestions for the next sentence in a story from a large corpus. Other methods proposed GUI and tools to facilitate co-creative narrative generation~\cite{manjavacas-etal-2017-synthetic,kapadia}.
Unlike us, these approaches explore the value of and tools for interaction rather than designing methods for incorporating user input into the model.

Another line of research decomposes story generation into two steps: story plot planning and plot-to-surface generation. Previous work 
produces story-plans based on sequences of events~\cite{Martin:18,ijcai2019-829, DBLP:conf/aaai/AmmanabroluTCLM20}, critical phrases~\cite{xu:18} or both events and entities ~\cite{fan:19}. ~\citeauthor{Yao:19} \shortcite{Yao:19} model the story-plan as a sequence of keywords. They proposed \emph{Static} and \emph{Dynamic} paradigms that generate a story based on these story-plans.~\citeauthor{Goldfarb-Tarrant:19}~\shortcite{Goldfarb-Tarrant:19} adopted the \emph{static} model proposed in \citeauthor{Yao:19} \shortcite{Yao:19} to supervise story-writing.

A major focus of these works is on generating 
a coherent plan for generating the story. In contrast, our contribution is complementary since we do not focus on \textit{planning} but on \emph{generation}. We present approaches to effectively incorporate external knowledge in the form of cue-phrases during generation, and conduct extensive experiments to compare our models with those of \citeauthor{Yao:19} \shortcite{Yao:19} by modifying them to work in our setup.

\begin{figure}[!t]
	\begin{center}
	\includegraphics[scale=0.52]{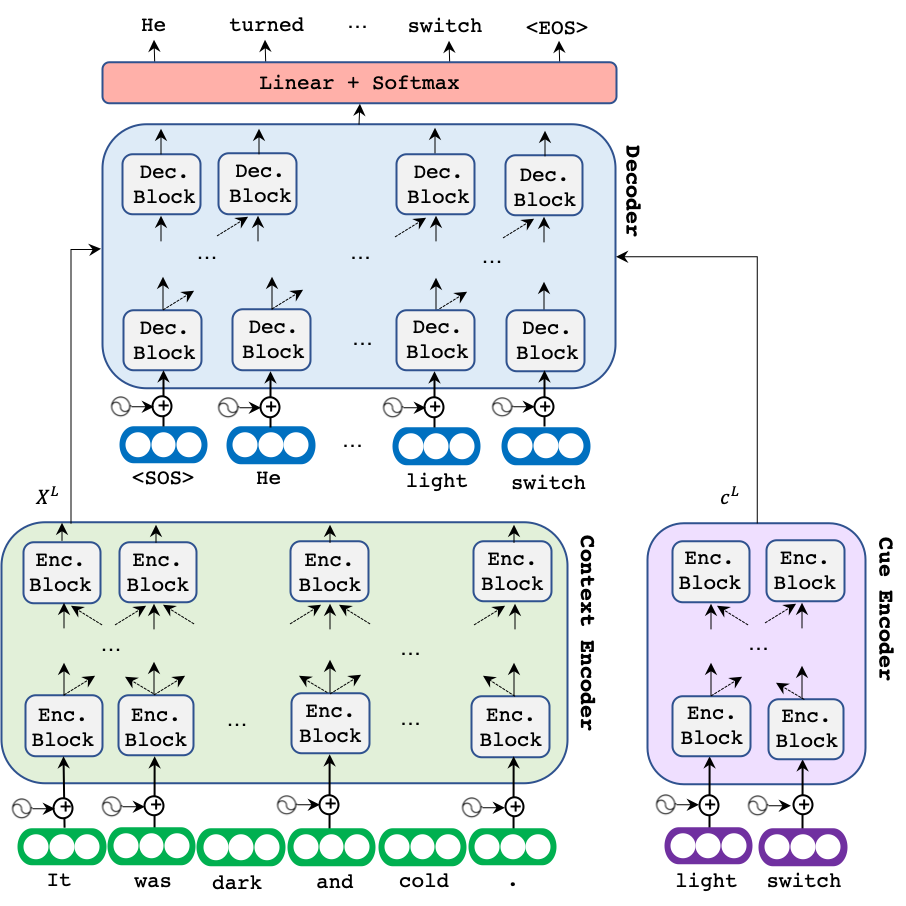}

	\caption{Overall model architecture for \textit{Cued
Writer} and \textit{Relevance Cued Writer}.}
	\label{modelarc}
	\end{center}
\end{figure}

\section{Interactive Story Generation}

We design models to generate a story one sentence at a time. Given the 
generated context  so far (as a sequence of tokens) $X=\{x_{1},...,x_{T}\}$, and the cue phrase for the next sentence $c=\{c_{1},...,c_{K}\}$, our models generate the tokens of the next sentence of the story $Y=\{y_{1},...,y_{M}\}$. We train the models by minimizing the cross-entropy loss:

\vspace{-0.4cm}
\begin{equation}
L_{\theta} = - \sum _{i=1}^{M} \log P(y_{i}|X,c, \theta)
\end{equation}

Here, $\theta$ refers to model parameters.
Note that when generating the $n$-th sentence, the model takes the first $n-1$ sentences in the story as the context along with the cue phrase.

In the rest of this section, we describe our two novel content-inducing approaches for addressing the interactive story generation task: the \textit{Cued Writer}, and the \textit{Relevance Cued Writer}. These models share an overall encoder-decoder based architecture shown in Fig.~\ref{modelarc}. They adopt a dual encoding approach where two separate but architecturally similar encoders are used for encoding the context (Context Encoder represented in the green box) and the cue phrase (Cue Encoder represented in the purple box). Both these encoders advise the Decoder (represented in the blue box), which in turn generates the next sentence. The two proposed models use the same encoding mechanism (described in \S~\ref{transbackground}) and differ only in their decoders (described in \S~\ref{incorpcue}).

\begin{figure}[t]
	\begin{center}
	\includegraphics[scale=0.41]{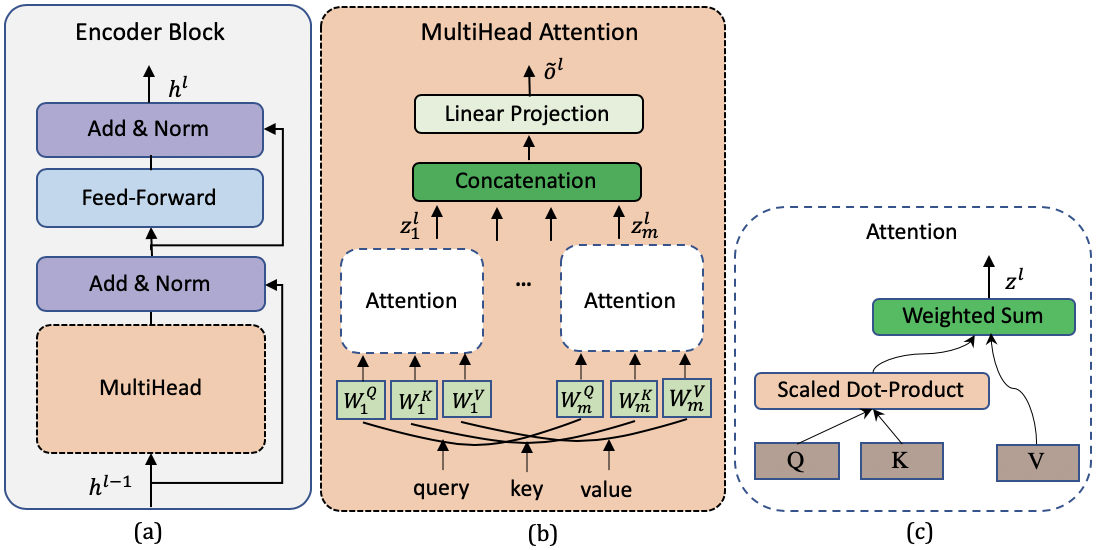}
	\caption{ \textbf{(a)} Encoder Block consists of MultiHead and FFN. \textbf{(b)} MultiHead Attention. \textbf{(c)} Attention Module.}
	\label{figure:encoderattn}
	\end{center}
\end{figure}

\subsection{Encoder \label{transbackground}} 
Our models use the Transformer encoder introduced in~\citeauthor{Vaswani:17} \shortcite{Vaswani:17}. Here, we  provide a generic description of the encoder architecture followed by the inputs to this architecture for the Context and Cue Encoders in our models.  

Each encoder layer $l$ contains architecturally identical Encoder Blocks, referred to as \textsc{EncBlock} (with unique trainable parameters). Fig.~\ref{figure:encoderattn}(a) shows an Encoder Block which consists 
of a Multi-Head attention and an FFN that applies the following operations: 

\vspace{-0.3cm}
\begin{subequations}
\begin{gather}
\Tilde{o}^l = \mathrm{M\pnt{ULTI}H\pnt{EAD}}(h^{l-1}) \\
o^l = \mathrm{L\pnt{AYER}N\pnt{ORM}}(\Tilde{o}^l + h^{l-1})\\
\Tilde{h}^l = \mathrm{FFN}(o^l)\\
h^l = \mathrm{L\pnt{AYER}N\pnt{ORM}}(\Tilde{h}^l + o^{l})
\end{gather}
\end{subequations}

Where \textsc{MultiHead} represents Multi-Head Attention (described below), FFN is a feed-forward neural network with ReLU activation~\cite{lecunn:15}, and \textsc{LayerNorm} is a layer normalization~\cite{Ba:16}. In the rest of the paper, \textsc{LayerNorm} (also shown as \texttt{Add \& Norm} in figures) is always applied after \textsc{MultiHead} and FFN, but we do not explicitly mention that in text or equations for simplicity.

\vspace{0.05cm}
\noindent\textbf{Multi-Head Attention} \space\space The multi-head attention, shown in Fig.~\ref{figure:encoderattn}(b), is similar to that used in~\citeauthor{Vaswani:17} \shortcite{Vaswani:17}. It is made of multiple Attention heads, shown in Fig.~\ref{figure:encoderattn}(c). The Attention head has three types of inputs: the query sequence, $Q \in R^{n_{q}\times d_k}$, the key sequence, $K\in R^{n_{k}\times d_k}$, and the value sequence, $V\in R^{n_{v}\times d_k}$. The attention module takes each token in the query sequence and attends to tokens in the key sequence using a scaled dot product. The score for each token in the key sequence is then multiplied by the corresponding value vector to form a weighted sum:
\begin{equation}
\mathrm{A\pnt{TTN}(Q,K,V) = softmax} \left( \frac{Q K^T}{\sqrt{d_k}} \right) V \normalsize
\label{eq1}
\end{equation}

\noindent For each head, all $Q$, $K$, and $V$ are passed through a head-specific projection prior to the attention being computed. The output of a single head is:
\begin{equation}
    H_i = \mathrm{A\pnt{TTN}}(QW_i^Q, KW_i^K, VW_i^V)
\end{equation}

Where $W$s are head-specific projections. Attention heads $H_i$ are then concatenated:
\begin{equation}
    \mathrm{M\pnt{ULTI}H(Q,K,V)} = [H_i;...;H_m]W^O
    \label{eq4}
\end{equation}

Where $W^O$ is an output projection. In the encoder, all query, key, and value come from the previous layer and thus:
\begin{equation}
    \mathrm{M\pnt{ULTI}H\pnt{EAD}}(h^{l-1}) = \mathrm{M\pnt{ULTI}H}(h^{l-1},h^{l-1},h^{l-1})
    \label{mul}
\end{equation}

\begin{figure}[t]
\centering
\vspace{-0.3cm}

\subfloat[][\scriptsize Cued Writer]{
    \includegraphics[width=0.4\textwidth]{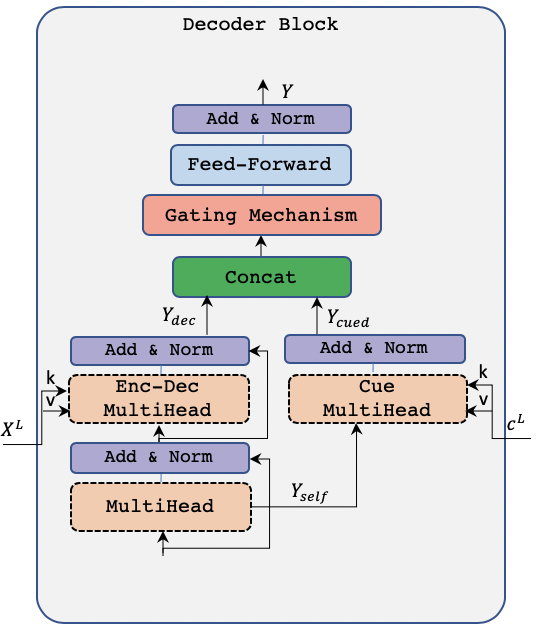}{ 
    \label{fig4a}}
    } \\
    \vspace{-8pt}
\subfloat[][\scriptsize Rel. Cued Writer]{ %
    \includegraphics[width=0.50\textwidth]{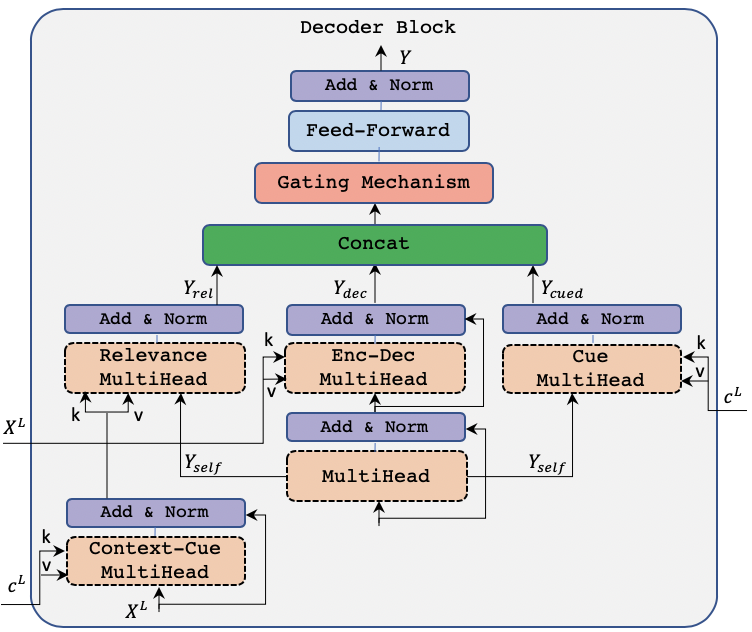}{
    \label{fig4b}} 
    }
    \vspace{-5pt}
\caption{Decoder architectures.  $X^L$ and $c^L$ are the outputs of the top-layers of the Context and Cue encoders respectively, and $K$ and $V$ are the corresponding keys and values.
}
\label{decoder}
\end{figure}

\noindent\textbf{Encoder Input} \space\space The Encoder Blocks described above form the constituent units of the Context and Cue Encoders, which process the context and cue phrase respectively. Each token in the context, $x_i$, and cue phrase, $c_i$, is assigned two kinds of embeddings: \emph{token embeddings} indicating the meaning and \emph{position embeddings} indicating the position of each token within the sequence. These two are summed to obtain individual input vectors, $X^0$, and $c^0$, which are then fed to the first layer of Context and Cue encoders, respectively. Thereafter, new representations are constructed through layers of encoder blocks: 
\begin{subequations}
\begin{gather}
\textbf{$X^{l+1}$} = \mathrm{E\pnt{NC}B\pnt{LOCK}}(X^l,X^l,X^l) \label{selfattn}\\
\textbf{$c^{l+1}$} = \mathrm{E\pnt{NC}B\pnt{LOCK}}(c^l,c^l,c^l) \label{selfattncue}
\end{gather}
\end{subequations}
where $l\in[0,L-1]$ denotes different layers. In Eqn. \ref{selfattn} and~\ref{selfattncue}, the output of the previous layer's Encoder Block is used as $Q$, $K$, and $V$ input for the multi-head attention of the next block.

\subsection{Content-Inducing Decoders}\label{incorpcue}
We now describe the decoders for our models. \par 
\noindent\textbf{Cued Writer} \space\space\space The main intuition behind our first model, \textit{Cued Writer}, is that since cue phrases indicate users' expectations of what they want to see in the next sentence of the story, they should be used by the model \emph{at the time of generation}, i.e., in the decoder. Below, we describe the decoder used by the \textit{Cued Writer}.

After processing the two types of inputs in the Context 
and Cue Encoders, the model includes their final encoded representations ($X^L$ and $c^L$) in the decoder. The decoder consists of $L$  layers with architecturally identical Decoder Blocks. Each Decoder Block contains \texttt{Enc-Dec MultiHead} and the \texttt{Cue MultiHead} units (see Fig. \ref{decoder}(a)), which let the decoder to focus on the relevant parts of the context and the cue phrase, respectively. 

Given $Y^0$ as the word-level embedding representation for the output sentence, our Decoder Block is formulated as:
\vspace{-0.2cm}
\begin{subequations}
\begin{gather}
\textbf{$Y_{self}^{l+1}$} = \mathrm{M\pnt{ULTI}H}(Y^l,Y^l,Y^l) \label{dec-self-attn}\\
\textbf{$Y_{dec}^{l+1}$} = \mathrm{M\pnt{ULTI}H}(Y_{self}^{l+1},X^{L},X^{L}) \label{enc-dec-attn}\\
\textbf{$Y_{cued}^{l+1}$} = \mathrm{M\pnt{ULTI}H}(Y_{self}^{l+1},c^{L},c^{L}) \label{cue-enc-dec-attn}
\end{gather}
\end{subequations}

Eqn.~\ref{dec-self-attn} is standard self-attention, which measures the intra-sentence agreement for the output sentence and corresponds to the \texttt{MultiHead} unit in Fig.~\ref{decoder}(a). Eqn.~\ref{enc-dec-attn}, describing the \texttt{Enc-Dec MultiHead} unit, measures the agreement between context and output sentence, where queries come from the decoder Multi-Head unit ($Y_{self}$), and the keys and values come from the top layer of the context encoder ($X^L$). Similarly, Eqn.~\ref{cue-enc-dec-attn} captures the agreement between output sentence and cue phrase through \texttt{Cue MultiHead} unit. Here, keys and values come from the top layer of the Cue encoder ($c^L$). \par 
Lastly, we adapt a gating mechanism~\cite{sriram2017cold} to integrate the semantic representations from both $Y_{dec}$ and $Y_{cued}$ and pass the resulting output to FFN function:
\begin{subequations}
\begin{gather}
\textbf{$g^{l+1}$} = \sigma (W_1[Y_{dec}^{l+1}; Y_{cued}^{l+1}]) \\
\textbf{$Y_{int}^{l+1}$} = W_2(g^{l+1} \circ [Y_{dec}^{l+1}; Y_{cued}^{l+1}])\\
\textbf{$Y^{l+1}$} = \mathrm{FFN}(Y_{int}^{l+1})
\label{ffn_decoder}
\end{gather}
\end{subequations}

\noindent the representation from $Y_{dec}$ and $Y_{cued}$ are concatenated to learn gates, $g$. The gated hidden layers are combined by concatenation and followed by a linear projection with the weight matrix $W_2$.

\noindent\textbf{Relevance Cued Writer} \space\space\space The decoder of \textit{Cued Writer} described above captures the relatedness of the context and the cue phrase to the generated sentence but does not study the relatedness or relevance of the cue phrase to the context.
We incorporate this relevance in the decoder of our next model, \textit{Relevance Cued Writer}. Its Decoder Block (shown in Fig.~\ref{decoder}(b)) is similar to that of \textit{Cued Writer} except for two additional units: the Context-Cue and Relevance MultiHead units. The intuition behind the \texttt{Context-Cue MultiHead} unit (Eqn.~\ref{cont-cue-rel}) is to characterize the relevance between the context and the cue phrase, so as to highlight the effect of words in the cue phrase that are more relevant to the context thereby promoting topicality and fluency. This relevance is then provided to the decoder using the \texttt{Relevance MultiHead} unit (Eqn.~\ref{rel-enc-dec}): 
\begin{subequations}
\begin{gather}
\textbf{$X_{rel}^{l+1}$} = \mathrm{M\pnt{ULTI}H}(X^{L}, c^{L}, c^{L}) \label{cont-cue-rel} \\
\textbf{$Y_{rel}^{l+1}$} = \mathrm{M\pnt{ULTI}H}(Y_{self}^{l+1}, X_{rel}^{l+1}, X_{rel}^{l+1}) \label{rel-enc-dec}
\end{gather}
\end{subequations}

\noindent We fuse the information from all three sources using a gating mechanism and pass the result to FFN:
\vspace{-0.2cm}
\begin{subequations}
\begin{gather}
\textbf{$g^{l+1}$} = \sigma (W_1[Y_{dec}^{l+1}; Y_{cued}^{l+1}; Y_{rel}^{l+1}]) \\
\textbf{$Y_{int}^{l+1}$} = W_2(g^{l+1} \circ [Y_{dec}^{l+1}; Y_{cued}^{l+1}; Y_{rel}^{l+1}])\\
\textbf{$Y^{l+1}$} = \mathrm{FFN}(Y_{int}^{l+1})
\end{gather}
\end{subequations}

Finally, for both models, a linear transformation and a softmax function (shown in Fig.~\ref{modelarc}) is applied to convert the output produced by the stack of decoders to predicted next-token probabilities:
\begin{equation}
P(y_{i}|y_{<i},X,c, \theta) = \mathrm{softmax}(Y_{i}^{L}W_y)
\end{equation}
\noindent where $P(y_{i}|y_{<i},X,c, \theta)$ is the likelihood of generating $y_i$ given the preceding text ($y_{<i}$), context and cue, and $W_y$ is the token embedding matrix.

\section{Empirical Evaluation \label{experiments}}

\subsection{Dataset \label{dataset}}
We used the ROCStories corpus ~\cite{Mos:16} for experiments. It contains $98,161$ five-sentence long stories with a rich set of causal/temporal sequences of events. 
We held out $10\%$ of stories for validation and $10\%$ for test set.

\begin{table*}[ht]
\centering
\setlength{\tabcolsep}{0.6em}
\footnotesize
\begin{tabular}{l|c c c c c c}
\toprule
 \textbf{Models} & PPL ($\downarrow$) & BLEU-1 ($\uparrow$) & BLEU-2 ($\uparrow$) & BLEU-3 ($\uparrow$) & GM ($\uparrow$) & Repetition-4 ($\downarrow$)\\ 
\midrule
  \textsc{Dynamic}~\cite{Yao:19} & 29.49 & 30.05 & 9.16 & 4.59 & 0.73 & 44.36\\
  \textsc{Static}~\cite{Yao:19} & 20.81 & 33.25 & 9.64 & 4.77 & 0.75 & 26.26\\\midrule
 \textsc{Seq2Seq}  & 20.97 & 33.91 & 10.01 & 3.09 & 0.82 & 33.23 \\
 \textsc{Vanilla} & 15.78 & 40.30 & 16.09 & 7.19 & 0.89 & 20.87\\  \midrule
Cued Writer & 14.80 & 41.50 & 16.72 & 7.25 & 0.92 & \textbf{15.08}\\
Rel. Cued Writer & \textbf{14.66} & \textbf{42.65} & \textbf{17.33} & \textbf{7.59} & \textbf{0.94} & 16.23\\
\bottomrule
\end{tabular}
\caption{\label{automatic} Automatic evaluation results. Our models outperform all baselines across all metrics ($p<0.05$).}
\vspace{-1em}
\end{table*}

\subsection{Baselines \label{bases}} 

\noindent \textbf{\textsc{Seq2Seq}} 
    \space\space Our first baseline is based on a LSTM sentence-to-sentence generator with attention~\cite{bahdanau:14}.  In order to incorporate user-provided cue phrases, we concatenate context and cue phrase with a delimiter token ($<$\$$>$) before passing it to the encoder. 

\noindent \textbf{\textsc{Dynamic}}  \space\space This is the Dynamic model proposed by~\citeauthor{Yao:19}~\shortcite{Yao:19} modified to work in our setting. For a fair comparison, instead of generating a plan, we provide the model with cue phrases and generate the story one sentence at a time. 
\par 

\noindent \textbf{\textsc{Static}}\space\space The \textsc{Static} model~\cite{Yao:19} gets all cue phrases at once to generate the entire story\footnote{We used the implementation available at: \bulurl{https://bitbucket.org/VioletPeng/language-model/}}. 
By design, it has additional access to all, including future, cue phrases. Our models and other baselines do not have this information.\par 
\noindent \textbf{\textsc{Vanilla}}\space\space To verify the effectiveness of our content-inducing approaches, we use a Vanilla Transformer as another baseline and concatenate context and cue phrase using a delimiter token.

\vspace{0.05cm}

\subsection{Training details \label{traindetail}}
Following previous work~\cite{Vaswani:17}, we initialize context encoders and decoders with $6$ layers ($512$ dimensional states and $8$ attention heads). Our models contain 3-layer encoders for encoding cue phrases (all other specifications are the same). For the position-wise feed-forward networks, we use $2048$ dimensional inner states. We use the Adam optimizer ~\cite{adam:14} with a learning rate of $0.0001$ and residual, embedding, and attention dropouts with a rate of $0.1$ for regularization. Models are implemented in PyTorch, trained for $30$ epochs with early stopping on validation loss.

\vspace{0.05cm}
\noindent\textbf{Cue-phrases for Training and Automatic Evaluation:} For training all models, we need cue phrases, which are, in principle, to be entered by a user. However, to scale model training, we automatically extracted cue phrases from the target sentences in the training set using the previously proposed RAKE algorithm~\cite{rose:10}. It is important to note that cue phrases can represent a variety of information, and many other methods can be used to extract them for training purposes. For example, 
topic words, 
distinctive entities or noun phrases in the sentence, 
the headword in the dependency parse of the sentence, etc.

Our automatic evaluations were done on a large-scale, and so we followed a similar approach for extracting cue-phrases.

\noindent \textbf{Cue-phrases for Human Evaluation:} In the interest of evaluating the interactive nature of our models, cue-phrases were provided manually during our interactive evaluations\footnote{We left the definition of cue-phrase open-ended to enable flexibility in user interaction. They are typically 1-2 words.}.

\noindent \textbf{General Statistics on Cue-phrases:}  Automatically extracted cue phrases has the vocabulary size of $22,097$, and $6,189$ on the train and test set, respectively with the average $10$\% coverage over the entire target sentence.
Cue-phrases are typically 1-2 words. Comparing user-provided vs automatically extracted cue-phrases, the average length of user-provided cue-phrases in interactive evaluation is $1.56$, with a vocabulary size of $206$, whereas these numbers are $1.59$ and $214$ for their corresponding automatically extracted cue phrases.

\subsection{Automatic Evaluation \label{automaticEval}}

Following previous credible works~\cite{Martin:18,Fan:18}, we compare various methods using Perplexity and BLEU~\cite{papineni2002bleu} on the test set. We reported BLEU-n for $n{=}1,2,3$. From Table~\ref{automatic}, we can see that both our models outperform 
\textsc{Dynamic} and \textsc{Static} by large margins on perplexity and BLEU scores. The proposed models are also superior to the \textsc{Seq2Seq} and \textsc{Vanilla} baseline on both measures. Comparing the last two rows of Table~\ref{automatic}, we also see an additive gain from modeling the relevance in  \textit{Rel. Cued Writer}. All improvements are statistically significant (approximate randomization~\cite{noreen:89}, $p<0.05$).\par 

To evaluate how well the story generation model incorporates the cues, we use an embedding-based greedy matching score (GM)~\cite{liu-etal-2016-evaluate}. The score measures the relatedness of the generated story with cues by greedily matching them with each token in a story 
based on the cosine similarity of their word embeddings~\cite{Yao:19}. We can see from the $5$th column in Table~\ref{automatic} that our models generate stories that are more related to the cue phrases.

\begin{figure}[t]
\centering
\vspace{-0.3cm}

\subfloat{ 
    \includegraphics[width=0.25\textwidth]{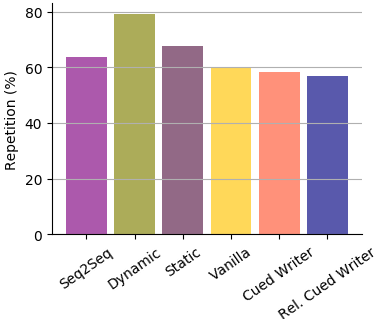}{
    \label{fig4a}}
    }
\subfloat{ 
    \includegraphics[width=0.25\textwidth]{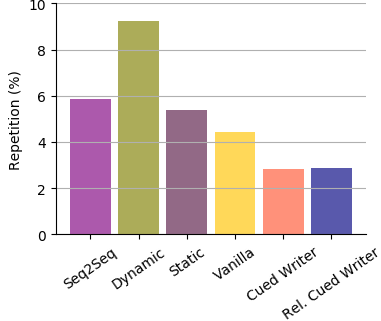}{
    \label{fig4b}}
    }
\caption{Inter-story (left) and Intra-story (right) repetition scores. The proposed models have better scores.
}
\label{rept}
\end{figure}

Previous works have shown that neural generation models suffer from repetition issue; and so we additionally evaluate the models using repetition-4 which measures the percentage of generated stories that repeat at least one 4-gram~\cite{Shao2019Planning} and inter- and intra-story repetition scores~\cite{Yao:19}. A lower value is better for these scores. The result of repetition-4 is reported in the last column of Table~\ref{automatic}. The proposed models significantly outperform all baselines, and among the two \textit{Cued Writer} is better. 
Inter and intra repetition scores are depicted in Fig.~\ref{rept}. Our two proposed models are almost comparable on these metrics but they show a general superior performance compared to all baselines. In particular, \textit{Rel. Cued Writer} achieves a significant performance increase of $16\%$ and $46\%$ on these scores 
over the stronger model of \citeauthor{Yao:19}~\shortcite{Yao:19}\footnote{Note that the result of our \textsc{Seq2Seq} baseline is not directly comparable with that of Inc-S2S in \cite{Yao:19}, since we included cue phrases as additional input whereas Inc-S2S generate the whole story conditioned on the title.}.

\begin{table}[t]
\scriptsize
	\begin{center}
\begin{tabular}{p{7.3cm}}
	\toprule[0.9pt]
	\textbf{Prompt (first sentence): } Jordan was watching TV on her couch. \\ \midrule
	\textbf{Cue phrases: }watch football - change channel - comedy show - very funny \\[0.2cm]
	She was trying to watch football on TV. Then she went to change channel. Finally, she decided to watch a comedy
show. She saw the comedy that was playing and didn't like.\\\midrule
	\textbf{Cue phrases: }soccer - cook - order pizza - tasty dinner \\[0.2cm]
	Her brother was playing in a soccer. She wasn't able to cook. Instead, she ordered pizza. Her brother was happy with the tasty dinner. \\
\bottomrule[0.9pt]
	
\end{tabular}
\vspace{-1em}
	\end{center}
\caption{\label{interc_eval} Example of stories generated in interactive evaluation using two models given the same prompt and different set of cue-phrase.} 
	\vspace{-1em}
\end{table}

\subsection{Human Evaluation } \label{humanEval}
Automatic metrics cannot evaluate all aspects of open-ended text generation~\cite{Fan:18}, and so we also conduct several human evaluations.

\noindent\textbf{Interactive Evaluation}
\space\space\space In this experiment, human subjects compare our best model, \textit{Rel. Cued Writer}, with the strongest baseline from the automatic evaluations (\textsc{Vanilla}) in an interactive, real-time setup.\par 
For robust evaluation, it is essential that the users generate a wide variety of stories. Since generating different prompts (first sentence) requires creativity on the part of human judges and can be challenging, we provided participants with initial prompts that were randomly selected from the test set. For each prompt, the participants generated stories using both models 
by interactively providing cue-phrases
\footnote{We included instructions and examples for participants. The order of the presentation of the models was random. The judges were 
self-identified native English speakers.}. They were then asked to choose which story they prefer. Participants preferred stories generated by \textit{Rel. Cued Writer} over \textsc{Vanilla} in $57.5\%$ of the cases ($80$ stories in total, $p\sim 0.1$).

\begin{figure}[t]
\centering

\subfloat[][]{ 
    \includegraphics[width=0.25\textwidth]{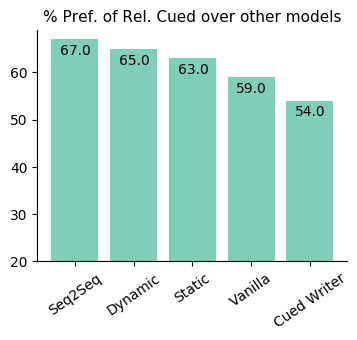}{ 
    \label{fig4a}}
    } 
\subfloat[][]{ 
    \includegraphics[width=0.23\textwidth]{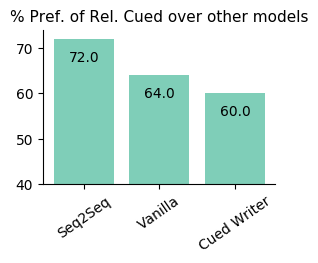}{
    \label{fig4b}} 
    }
    \vspace{-8pt}
\caption{Human evaluations on story-level (left) and sentence-level (right). We find that human judges preferred stories generated by Rel. Cued Writer.}
\label{AMTres}
\end{figure}

Judges also rated the stories in terms of fluency and coherence on a 5-point Likert scale.
\textit{Rel. Cued Writer} achieved a higher fluency score of $4.22$ compared with $3.80$ achieved by \textsc{Vanilla}. \textsc{Vanilla} attained a slightly higher coherence score ($3.40$ vs. $3.35$). On manually inspecting the generated stories, we found that our model generates longer sentences (avg. $9.18$ words) with more complex language, whereas \textsc{Vanilla} generated relatively shorter sentences (avg. $7.46$ words) which might improve coherence.\par
This experiment is promising but inconclusive because for the same prompt, the participants could provide different sets of cue-phrases for different models, resulting in generated stories that are too different to be comparable (Table~\ref{interc_eval} shows an example). This led us to conduct the following more controlled evaluations.

\begin{table*}[!ht]
\scriptsize
\def\arraystretch{1.5}
	 \begin{tabular}{p{1.75cm}p{13.5cm}} 
	\toprule[0.9pt]
	\textbf{Prompt} & I was coughing very badly. \space\space\space $\mid$ \space\space\space  \textbf{Cue phrases} \space\space\space notice blood - emergency room - waited - okay \\ \midrule\midrule 
	\textbf{\textsc{Dynamic}} & I noticed that I had a headache. I went to the doctor. The doctor told me that I had to go to  \textbf{\textcolor{red}{the doctor}}. The doctor said I was fine. \textcolor{violet}{\textbf{[waited]}} \\\midrule
	\textbf{\textsc{Static}} & \textbf{\textcolor{blue}{She}} noticed a huge blood pressure. She went to the doctor. \textbf{\textcolor{blue}{The doctor told her to stop.}} She waited for a while. \textbf{\textcolor{violet}{[okay]}} \\ \midrule 
	\textsc{\textbf{Seq2Seq}} &  I noticed blood and \textcolor{red}{\textbf{noticed blood}} in the closet. \textcolor{blue}{\textbf{Now I have emergency room}} for \textcolor{red}{\textbf{emergency room}}. I found it and I waited until the next day. I feel guilty and now i don't have to do it again. \textcolor{violet}{\textbf{[okay]}}\\ \midrule 
	\textsc{\textbf{Vanilla}} & I noticed blood in the bathroom. I had to go to the emergency room. I waited in the bathroom for a while. Luckily \textbf{\textcolor{blue}{the doctor was  okay.}}  \\ \midrule 
	\textbf{Cued Writer} & I noticed blood in my mouth. I had to go to the emergency room. I waited there a few hours. I felt okay.\\ \midrule
	\textbf{Rel Cued Writer} & I noticed blood on my shirt. I went to the emergency room. I waited for the doctor. He came and I was okay.\\
	\toprule[0.9pt]
	
	\textbf{Prompt} & Harry really wanted to see the new Star Wars movie. \space\space\space $\mid$ \space\space\space  \textbf{Cue phrases} \space\space\space opening day - ticket - entire movie - really \\ \midrule\midrule
	\textbf{\textsc{Dynamic}} & He went to the theater to show up. He bought a ticket. He had a great time. \textcolor{red}{\textbf{He had a great time.}}  \textcolor{violet}{\textbf{[opening day] [entire movie]}} \\\midrule
	\textbf{\textsc{Static}} &  One day, \textcolor{blue}{\textbf{she}} was in a hurry to get to the theater. she wasn't sure if she would like to go. She was excited to try the new movie. \textcolor{red}{\textbf{She was so excited}} to see the movie. \textbf{\textcolor{violet}{[opening day] [ticket]}} \\ \midrule
	\textsc{\textbf{Seq2Seq}} &  The day \textcolor{red}{\textbf{day}} was opening day and his family was \textcolor{red}{\textbf{opening}} the \textcolor{red}{\textbf{opening day}}. Harry had bought a ticket and the \textcolor{red}{\textbf{ticket}} wasn't very good. The entire movie was very happy. Harry became \textcolor{blue}{\textbf{very really}} disappointed. \\ \midrule
	\textsc{\textbf{Vanilla}} & On opening day, Harry was very nervous. He bought a ticket \textcolor{blue}{\textbf{to}} the theater. He bought \textcolor{blue}{\textbf{Harry}} ticket \textcolor{red}{\textbf{tickets}} to the theater. He really didn't like the movie. \textcolor{violet}{\textbf{[entire movie]}} \\\midrule
	\textbf{Cued Writer} & On opening day, he went to the theater . He bought a ticket at the theater. The entire movie was great. He really was excited. \\\midrule
	\textbf{Rel Cued Writer} & He decided to watch it on opening day. He got to the theater and got a ticket. He watched the entire movie. He was really excited about it.\\\midrule
	
\end{tabular}
	\caption{\label{samples} Sample stories generated by different models. We highlight in different color the \textbf{\textcolor{violet}{[missing]}} cue phrase, \textbf{\textcolor{blue}{incoherent or unfluent}}, and \textbf{\textcolor{red}{repetitive}} parts of each story. We see that compared to baselines, our models correctly mention cue phrases and generate better stories.}
	\vspace{-1em}
\end{table*}

\noindent\textbf{Story-level Evaluation} \space\space\space In this experiment, we again make pairwise comparisons, but both models are provided the same prompts, and sets of cue phrases\footnote{Cue-phrases were extracted automatically to alleviate any biases, then curated manually.}.
$3$ judges evaluated $100$ pairs of stories (in shuffled order)\footnote{For all AMT evaluations, we 
only allow judges 
who have at least $500$ previous jobs, and at least $97$\% approval rate.}. \par

Fig.~\ref{AMTres}(a) shows the percentage of preference for our stronger model, \textit{Rel. Cued Writer}, over the baselines. Judges prefer our model 
over all other baselines.
Also, judges preferred 
\textit{Rel. Cued Writer} over 
\textit{Cued Writer}, which demonstrates the effectiveness of the additional Context-Cue and Relevance Multi-Head units. All improvements are statistically significant (app. rand., $p<0.05$).

\noindent\textbf{Sentence-level Evaluation} \space\space\space
We also performed a more fine-grained evaluation of the models by evaluating generated sentences while the model is generating a story.
The generated sentences are evaluated 
in light of the (incomplete) story. Specifically, we provide an (incomplete) story passage and a manually provided cue phrase to the two models to generate the next sentence. We asked human judges to identify which of the two sentences is better based on their fluency and semantic relevance to (1) the input (incomplete) story and (2) the cue phrase. We did this experiment for a set of $100$  randomly selected stories ($400$ sentences. $3$ different judges evaluated each sentence pair. Fig.~\ref{AMTres}(b) shows that the \emph{Rel. Cued Writer} model was preferred over \textsc{Seq2Seq} and  \textsc{Vanilla} in $72\%$ and $64\%$ of the cases, respectively.  Comparing the two proposed models, we again see additive gain by modeling Cue-Context relevance. All improvements are statistically significant (app. rand., $p<0.001$).\par

\begin{table}[t]
\vspace{-0.25cm}
\footnotesize
\def\arraystretch{1.5}
	\begin{center}
\begin{tabular}{p{7.2cm}}
\toprule
	\textbf{off-topic: }Kelly and her friends went to a new ice-cream shop. They decided to try the new flavors. They all tried on many different restaurants. To their surprise, they thought it tasted good. They were glad to find one online.\\ 
	\textbf{Not-logically-consistent: } Avery received a homework assignment due in two weeks. He immediately read it. When he turned it in, he made schedule. He completed tasks and turned it in time. When he finished early, he was disappointed. \\
	\textbf{non-coreferent-pronouns: }Rob has never been on a rollercoaster. They go on all the way to six flags. He got on with a free ticket. Rob joined the rollercoaster. There was a long line of people in the line.\\ 
\bottomrule
	
\end{tabular}
	\end{center}
	\caption{\label{errors} Examples of errors made by our model.}
	\vspace{-1em}
\end{table}

\section{Qualitative Results and Error Analysis\label{qualitative} }

Table \ref{samples} presents examples of stories generated by different models for the same prompt and cue phrases. We highlight the \textcolor{violet}{[missing]} cue phrases, \textcolor{blue}{incoherent or unfluent}, and \textcolor{red}{repetitive} parts of each story. Note that we did not highlight \textcolor{violet}{[missing]}, if the model mentions part of the cue phrase or incorporates it semantically.
As we observe, all of the baselines suffered from 
several issues; however, our novel content inducing approaches generate more causally related sentences, which fit the given prompt and cue phrases more naturally.

We also manually reviewed $50$ stories, generated from our models 
and analyzed common errors. Table~\ref{errors} shows sample 
stories that depict different types of errors including ``getting off-topic'', ``not-logically-connected'' and ``non-coreferent pronouns''. The last type of error represents the cases where the model generates pronouns that do not refer to any previously mentioned entity. The examples demonstrate that there are still many challenges in this domain.

\section{Conclusion and Future Work}
This paper explored the problem of interactive storytelling, which leverages human and computer collaboration for creative language generation.
We presented two content-inducing approaches that take user-provided inputs as the story progresses and effectively incorporate them in the generated text.
Experimental results show that our methods outperform competitive baselines.
However, 
there are several other significant aspects to be considered in story generation, such as 
modeling of discourse relations, and representation of key narrative elements, which lie beyond the scope of this investigation. Also, while we received encouraging feedback from users on this setup during the interactive evaluation, we did not explore important questions about user interfaces, design, and human computer interaction. Future work can explore these questions and also explore other forms of natural language interaction.

\bibliographystyle{acl_natbib}
\bibliography{acl2020}

\end{document}